\documentclass[10pt,twocolumn,letterpaper]{article}

\usepackage{iccv}
\usepackage{times}
\usepackage{epsfig}
\usepackage{graphicx}
\usepackage{amsmath}
\usepackage{amssymb}

\usepackage{verbatim}
\usepackage[linesnumbered,lined,ruled]{algorithm2e}
\usepackage{bm}
\usepackage{enumerate}
\usepackage{mathtools}
\usepackage{threeparttable}
\usepackage{graphics}
\usepackage{subfig}
\usepackage{color}
\usepackage{booktabs}
\usepackage{appendix}

\usepackage[pagebackref=true,breaklinks=true,letterpaper=true,colorlinks,bookmarks=false]{hyperref}

\iccvfinalcopy 

\def\iccvPaperID{1125} 

\ificcvfinal\fi
\begin{document}

\title{EAT-NAS: Elastic Architecture Transfer\\for Accelerating Large-scale Neural Architecture Search}

\author{
Jiemin Fang$^{1*\dagger}$, Yukang Chen$^{3\dagger}$, Xinbang Zhang$^3$, Qian Zhang$^{2}$\\ 
Chang Huang$^{2}$, Gaofeng Meng$^3$, Wenyu Liu$^{1}$, Xinggang Wang$^{1}$ \\
$^1$\emph{School of EIC, Huazhong University of Science and Technology} $\; ^2$\emph{Horizon Robotics}\\
$^3$\emph{National Laboratory of Pattern Recognition, Institute of Automation, Chinese Academy of Sciences}\\
\emph{\{jaminfong, xgwang, liuwy\}@hust.edu.cn}$\;$
\emph{\{qian01.zhang, chang.huang\}@horizon.ai}\\
\emph{\{yukang.chen, xinbang.zhang, gfmeng\}@nlpr.ia.ac.cn}
}

\maketitle

\begin{abstract}
Neural architecture search (NAS) methods have been proposed to release human experts from tedious architecture engineering. However, most current methods are constrained in small-scale search due to the issue of computational resources. Meanwhile, directly applying architectures searched on small datasets to large datasets often bears no performance guarantee. This limitation impedes the wide use of NAS on large-scale tasks. To overcome this obstacle, we propose an elastic architecture transfer mechanism for accelerating large-scale neural architecture search (EAT-NAS). In our implementations, architectures are first searched on a small dataset, \eg, CIFAR-10. The best one is chosen as the basic architecture. The  search process on the large dataset, \eg, ImageNet, is initialized with the basic architecture as the seed. The large-scale search process is accelerated with the help of the basic architecture. What we propose is not only a NAS method but a mechanism for architecture-level transfer.

In our experiments, we obtain two final models EATNet-A and EATNet-B that achieve competitive accuracies, 74.7\% and 74.2\% on ImageNet, respectively, which also surpass the models searched from scratch on ImageNet under the same settings. For the computational cost, EAT-NAS takes only less than 5 days on 8 TITAN X GPUs, which is significantly less than the computational consumption of the state-of-the-art large-scale NAS methods.
\footnote[1]{Our pretrained models and the evaluation code are released at \url{https://github.com/JaminFong/EAT-NAS}}
\let\thefootnote\relax\footnote[0]{$^*$ The work was done during an internship at Horizon Robotics.}
\let\thefootnote\relax\footnote[0]{$^\dagger$ Equal contributions.}

\end{abstract}

\vspace{-10pt}
\section{Introduction}
Designing neural network architectures by human experts often requires tedious trials and errors. To make this process more efficient, many neural architecture search (NAS) methods~\cite{zoph2017learning,Real2018Regularized,pham2018efficient} have been proposed. Despite their remarkable results, most NAS methods require expensive computational resources. For example, 800 GPUs across 28 days are used by NAS~\cite{zoph2016neural} on the task of CIFAR-10~\cite{krizhevsky2009learning} image classification. Real-world applications involve lots of large-scale datasets. However, directly carrying out the architecture search on large-scale datasets, \eg, ImageNet~\cite{DBLP:conf/cvpr/DengDSLL009}, requires much more computation cost which limits the wide application of NAS. Although some accelerating methods have been proposed~\cite{zoph2017learning,Real2018Regularized,pham2018efficient}, few of them directly explore on large-scale tasks.

From lots of previous works, \eg VGGNet~\cite{simonyan2014very}, Goog-LeNet~\cite{DBLP:conf/cvpr/SzegedyLJSRAEVR15}, ResNet~\cite{he2016deep}, \etc, a neural architecture that has good performance on one dataset usually performs well on other datasets or tasks. PNAS~\cite{liu2017progressive} suggests the transfer capability of the searched architectures by measuring the correlation between performance on CIFAR-10 and ImageNet for different neural architectures.
Most existing NAS methods~\cite{zoph2017learning,Real2018Regularized,liu2017progressive} search for architectures on a small dataset, \eg, CIFAR-10~\cite{krizhevsky2009learning}, and then apply these architectures directly on a large dataset, \eg, ImageNet, with the architectures adjusted manually. Normally, when transferring to the large dataset, both the number of stacked cells / layers and filters in the network will be enlarged. The searched cell structures or layer operations remain unchanged. 
However, due to the dataset bias~\cite{tommasi2017deeper} between the small dataset and the large dataset, the best representation of the neural architecture differs between datasets. Moreover, because of the lower resolution and the limited number of training images and data categories in the small dataset, the effectiveness of the architecture or the cells / layers of it searched on the small dataset degrades, when being directly applied on the large dataset or transferred with only the depth and width adjusted by handcraft.

\begin{figure*}[ht]
    \setlength{\abovecaptionskip}{-5pt} 
    \setlength{\belowcaptionskip}{-15pt}
        \begin{center}
            \includegraphics[width=0.9\linewidth]{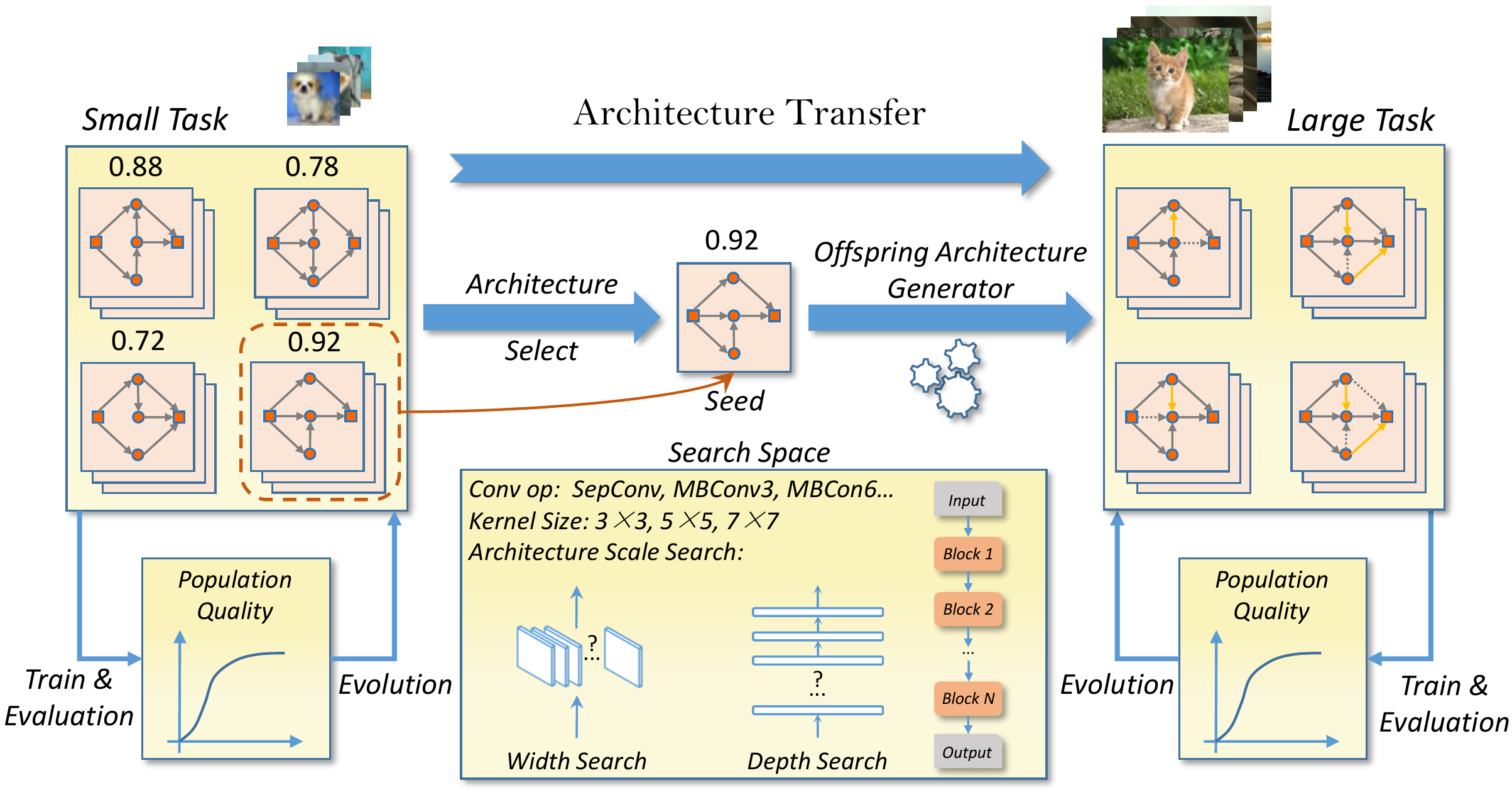}
        \end{center}
        \caption{The framework of Elastic Architecture Transfer for Neural Architecture Search (EAT-NAS). We firstly search for the basic architecture on the small-scale task and then search on the large-scale task with the basic architecture as the seed of the new population initialization.
        }
        \label{fig:EAT-framework}
\end{figure*}    

In this work, we propose a transfer learning solution to the above problem. What we propose is not only a NAS method but a common mechanism for architecture-level transfer learning. We define the elements in the neural architecture design as \emph{architecture primitives}, \eg the number of filters, the number of layers, operation types, the connection mode, kernel sizes. Our proposed elastic architecture transfer (EAT) is a mechanism which automatically transfers the neural architecture to a large-scale dataset, as is shown in Fig.~\ref{fig:EAT-framework}. It is elastic because all the architecture primitives are adjusted or fine-tuned automatically when transferring the architecture to another dataset.

In our implementation, we choose \emph{Evolutionary Algorithm} (EA) to design the basic NAS method. Benefitting from the population-based search process, the information of the architecture searched on the small dataset is easy to transfer to another architecture population on the large dataset. We adopt a two-stage search process to achieve the architecture transfer between different datasets. Firstly, we carry out the first search stage on a small dataset. When the first search process is finished, we choose the best architecture in the population as the seed which is called \emph{basic architecture}. Secondly, we use the seed to initialize the architecture population of the second search stage on the large dataset. We obtain the new architectures of the new population by adding some perturbations to the basic architecture. Finally, we carry on with the search process on the large dataset for fewer epochs. The population resulting from the deformed representations of the basic architecture can evolve faster towards the direction that fits the new dataset.

EAT narrows the gap between datasets on the architecture level automatically. The search process on the large dataset is accelerated by taking advantage of the information from the architecture searched on the small dataset. Our proposed mechanism of the architecture-level transfer supplies a new aspect for transfer learning. The transfer mechanism could not only be deployed to the EA-based NAS method, but also the gradient or RL based ones. Moreover, EAT could be available for various tasks and datasets in practical problems.

Our contribution can be summarized as follows:
\vspace{-5pt}
\begin{itemize}
\setlength{\itemsep}{0pt}
\setlength{\parsep}{0pt}
\setlength{\parskip}{0pt}
    \item [1)] We propose an elastic architecture transfer mechanism which automates the architecture transfer between datasets and enables performing NAS on large datasets with less computation cost.
    \item [2)] Through the experiments on the large-scale dataset, \ie, ImageNet, we show the efficiency of our method by cutting the search cost to only less than 5 days on 8 TITAN X GPUs, about 106x lower than the cost for MnasNet estimated based on~\cite{2018arXiv180711626T}. 
    \item [3)] Our searched architectures achieve remarkable ImageNet performance that is comparable to MnasNet which searches directly on the full dataset with huge computational cost (74.7\% vs 74.0\%).
\end{itemize}

\section{Related Work and Background}

\subsection{Neural Architecture Search}
Generating neural architectures automatically has arous-ed great interests in recent years. In NAS~\cite{zoph2016neural}, an RNN network trained with reinforcement learning is utilized as a controller to determine the operation type, parameter, and connection for every layer in the architecture. Although NAS~\cite{zoph2016neural} achieves impressive results, the search process is incredibly computation hunger and hundreds of GPUs are required to generate a high-performance architecture on CIFAR-10 datasets. Based on the NAS method in \cite{zoph2016neural}, many novel methods have been proposed to improve the efficiency of architecture search like finding out the blocks of the architecture instead of the whole network~\cite{zoph2017learning,zhong2018practical}, progressive search with performance predictor~\cite{liu2017progressive}, early stopping strategy in \cite{zhong2018practical}, and parameter sharing in \cite{pham2018efficient}. Though they have achieved impressive results, the search process is still computation hunger and extremely hard when the searched datasets are in large-scale, \eg, ImageNet. 

Another stream of NAS works utilizes the evolutionary algorithm to generate coded architectures~\cite{miikkulainen2019evolving, Real2018Regularized, 2018arXiv181003522L}. Modifications to the architecture (filter sizes, layer numbers, and connections) serve as the mutation in the search process. Though they have achieved state-of-the-art results, the computation cost is also far beyond affordable.

Besides, gradient-based NAS methods~\cite{liu2018darts, xie2018snas, cai2018proxylessnas} become popular. Gradient-based methods discard the black-box searching method and introduce architecture parameters, which are updated on the validation set by gradient descent, for every path of the network. A \emph{softmax} classifier is utilized to select the path and the operation for each node. The search space is relaxed to be continuous so that the architecture can be optimized with respect to its validation set performance by gradient descent. Though gradient-based NAS methods achieve great performance with high efficiency, the search space relies on the super network. All the possible sub-architectures should be included in the delicately designed super network. This suppresses the expansibility of the search space, \eg, it is not easy to search for the width of the architecture by gradient-based methods. For our evolutionary algorithm based EAT method over the discrete search space, the architecture can be encoded and is easier to be extended to diverse search spaces.

Recently, MnasNet~\cite{2018arXiv180711626T} proposes to search directly on large-scale datasets with latency optimization of the architecture based on RL. MnasNet successfully generates high-performance architectures with promising inference speed, but it requires huge computational resources. In total, 8K models are sampled to be trained on the nearly whole training set for 5 epochs and evaluated on the 50K validation set. It takes about 91K GPU hours as is estimated according to the description in~\cite{2018arXiv180711626T}.

There is a work \cite{NIPS2018_8056} that combines transfer learning with RL-based NAS method. They transfer the controller by reloading the parameters of the pretrained controller and add a new randomly initialized embedding for the new task. Our proposed elastic architecture transfer method focuses on transferring on the architecture-level automatically. The architecture searched on the small dataset can be transferred to the large dataset fast and precisely. EAT-NAS obtain models with competitive performance and much less computational resources than search from scratch.

\subsection{Evolutionary Algorithm based NAS}

\setlength{\textfloatsep}{0pt} 
\begin{algorithm}[t]
    \caption{Evolutionary Algorithm} 
    \label{algo:ea}
    \SetKwInOut{Input}{input}\SetKwInOut{Output}{output}
    \Input{population size $P$, sample size $S$, dataset $D$}
    \Output{the best model $M_{best}$}
    \begin{small}
        \texttt{$\mathbb{P}^{(0)}$ $\gets$ initialize($P$)}
        
        \For{\emph{$j$ $<$ $P$}} {
            $M_j.acc$ \texttt{$\gets$ train-eval(}$M_j$, $D$\texttt{)}
            
            $M_j.score$ \texttt{$\gets$ comp-score(}$M_j, M_j.acc$\texttt{)}
            }
            
        $Q^{(0)}$ \texttt{$\gets$ comp-quality(}$\mathbb{P}^{(0)}$\texttt{)}
            
        \While {$Q^{(i)}$ not converge} {
            \texttt{$S^{(i)}$ $\gets$ sample($\mathbb{P}^{(i)}$, $S$)}
            
            \texttt{$M_{best}$,$M_{worst}$ $\gets$ pick($S^{(i)}$)}
            
            \texttt{$M_{mut}$ $\gets$ mutate($M_{best}$)}
            
            \texttt{$M_{mut}.acc$ $\gets$ train-eval($M_{mut}$, $D$)}
            
            \texttt{$M_{mut}.score$~$\gets$~comp-score($M_{mut}$,~$M_{mut}.acc$)} 
            
            \texttt{$\mathbb{P}^{(i+1)}$ $\gets$ remove $M_{worst}$ from $\mathbb{P}^{(i)}$}
            
            \texttt{$\mathbb{P}^{(i+1)}$ $\gets$ add $M_{mut}$ to $\mathbb{P}^{(i)}$}

            \texttt{$Q^{(i+1)}$ $\gets$ comp-quality($\mathbb{P}^{(i+1)}$)}

            \texttt{$i$ ++}
            }
            
            \texttt{$M_{best}$ $\gets$ rerank-topk($\mathbb{P}_{best}$, $k$)}
    \end{small}
\end{algorithm}
\setlength{\floatsep}{0pt}

Evolutionary algorithm (EA) is widely utilized in NAS~\cite{miikkulainen2019evolving, Real2018Regularized, 2018arXiv181003522L}. As is summarized in Algorithm~\ref{algo:ea}, the search process is based on the population of various models. Conventionally, the population $\mathbb{P}$ is first initialized with randomly generated $P$ models which are within the setting range of the search space. Each model is trained and evaluated on the dataset to get the accuracy of the model.

Following the Pareto-optimal problem~\cite{deb2014multi}, we use \(acc \times [size / T]^\omega\) to compute the score of the model, where $acc$ denotes the accuracy of the model, $size$ denotes the model size, \ie, the number of parameters or multiply-add operations, $T$ is the target model size and $\omega$ is a hyperparameter for controlling the trade-off between accuracy and model size.
At each evolution cycle, $S$ models are randomly sampled from the population. The model with the best score and the worst one are picked up. The mutated model is obtained by adding some transformation to the one with the best score. The mutated model is trained, evaluated and added to the population with its score. The worst model is removed meanwhile. The above search process is called \emph{tournament selection}~\cite{goldberg1991comparative}. Finally, the top-k performing models are retrained to select the best one. Our architecture search method is based on the evolutionary algorithm.

\label{subsec:ss}
\begin{figure*}[ht]
    \setlength{\abovecaptionskip}{-5pt} 
    \begin{center}
        \includegraphics[width=0.9\linewidth]{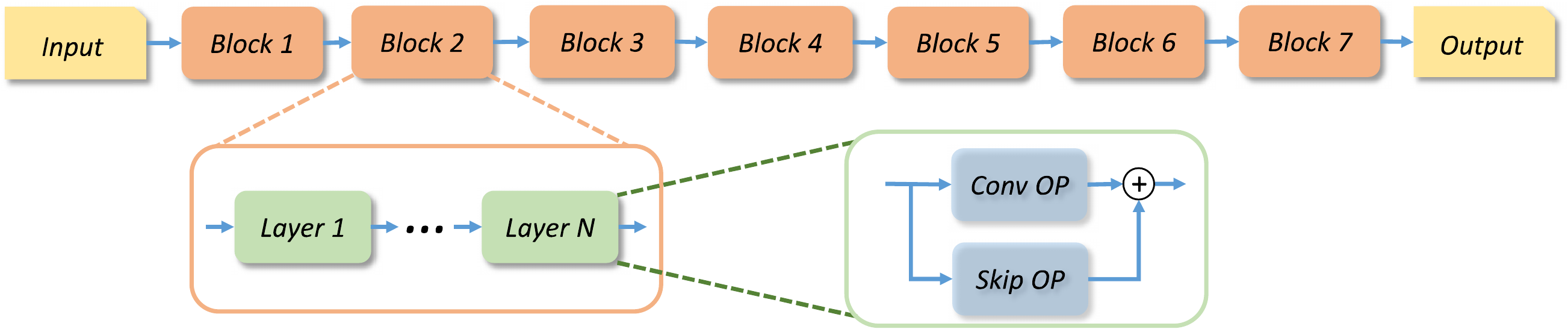}
    \end{center}
    \caption{Search Space. During search, all the blocks are concatenated to constitute the 
    whole network architecture. Each block consists of several layers and is represented by the following five primitives: convolutional operation type, kernel size, skip connection, width and depth.
    }
    \label{fig:search space}
\vspace{-15pt}
\end{figure*}

\section{Method}
To apply an architecture to large-scale tasks, most architecture search methods~\cite{Real2018Regularized,pham2018efficient,liu2017hierarchical, zoph2017learning} merely rely on prior knowledge of human experts. They transfer an architecture manually with only expanding the depth and width by multiplication or direct addition. Different from these conventional transfer methods, we propose an elastic architecture transfer (EAT) method. EAT automatically transfers the neural architecture to a large-scale task by fine-tuning the architecture primitives searched on a small-scale task. It is elastic for the transfer capability of all the architecture primitives, \eg, operator types, the structure, the depth and width of the architecture. EAT accelerates the large-scale search process by making use of the knowledge from the base architecture searched on the small-scale task. EAT adjusts the basic architecture to the large-scale task with all the architecture primitives fine-tuned.

\subsection{Framework}
Fig.~\ref{fig:EAT-framework} illustrates the process of EAT. The two search process on the small and the large datasets are based on the same \emph{search space}~(Section~\ref{subsec:ss}). We firstly search for a set of top-performing architectures on the small dataset, such as CIFAR-10. To get better performing architectures, we search for the \emph{architecture scale}~(Section~\ref{subsec:ass}) with the help of the width and depth factor. We design a criterion as \emph{population quality}~(Section~\ref{subsec:pop_q}) to better evaluate the model population. Then we retrain the top-performing models and choose the best one as the basic architecture.
Secondly, we start the architecture search on the large-scale task with the basic architecture as the seed to initialize the new architecture population. We design an \emph{architecture perturbation function}~(Section~\ref{APF}) to produce architectures of the new population. Then we continue the architecture search on the large-scale task based on the population derived from the basic architecture. In this way, the search on the new task is accelerated and obtains better performing models than carrying it out from scratch, which is benefited from the useful information of the basic architecture. Section~\ref{subsec:ablation-study} displays the results of contrast. Finally, we select the best one from the top-k performing models in the population by retraining them on the full large-scale dataset. Algorithm~\ref{algo:eat} displays the whole procedure for the elastic architecture transfer.

\begin{algorithm}[t]
    \caption{Elastic Architecture Transfer} 
    \label{algo:eat}
    \SetKwInOut{Input}{input}\SetKwInOut{Output}{output}
    \Input{datasets $D_1$, $D_2$, population size $P$}
    \Output{the target architecture $Arch_{target}$}
    \begin{small}
        \texttt{// initialize the population on $D_1$}

        $\mathbb{P}_1$ \texttt{$\gets$ initialize(}$P$\texttt{)}

        \texttt{evolve(}$\mathbb{P}_1$, $D_1$\texttt{)}

        $Arch_{basic}$ \texttt{$\gets$ rerank-topk \& select(}$\mathbb{P}_{1}$, $k$\texttt{)}

        \texttt{// initialize the population on $D_2$}

        \For{\emph{$i$ $<$ $P$}} 
        {
            $Arch_i$ \texttt{$\gets$ arch-perturbation(}$Arch_{basic}$\texttt{)}

            $\mathbb{P}_2$\texttt{.append(} $Arch_i$ \texttt{)}
        }

        \texttt{evolve(}$\mathbb{P}_2$, $D_2$\texttt{)}

        $Arch_{target}$ \texttt{$\gets$ rerank-topk \& select(}$\mathbb{P}_{2}$, $k$\texttt{)}

    \end{small}
\end{algorithm}

\subsection{Search Space}

A well-designed search space is essential for NAS. Inspired by MnasNet~\cite{2018arXiv180711626T}, we employ an architecture search space with MobileNetV2~\cite{sandler2018mobilenetv2} as the backbone. As Fig.~\ref{fig:search space} shows, the network is divided into several blocks which can be different from each other. Each block consists of several layers, whose operations are determined by a per-block sub search space. Specifically, the sub search space for one block could be parsed as follows:

\begin{itemize}
\vspace{-10pt}
\setlength{\itemsep}{0pt}
\setlength{\parsep}{0pt}
\setlength{\parskip}{0pt}
    \item \emph{Conv operation}: 
    depthwise separable convolution (SepConv)~\cite{chollet2017xception}, mobile inverted bottleneck convolution (MBConv) with
    diverse expansion ratios \{3,6\}~\cite{sandler2018mobilenetv2}.
    \item \emph{Kernel size}: 
    3$\times$3, 5$\times$5, 7$\times$7.
    \item \emph{Skip connection}: 
    whether to add a skip connection for every layer.
    \item \emph{Width factor}:
    the expansion ratio of the output width to the input width, $factor_{width} = N_{o} / N_{i}$, [0.5, 1.0, 1.5, 2.0].
    \item \emph{Depth factor}:
    the number of layers per block, [1, 2, 3, 4].
\vspace{-10pt}
\end{itemize}

Besides, the down-sampling and width expansion operations are applied in the first layer of each block.

To manipulate the neural architecture more conveniently, every architecture is encoded following the format defined in the search space. As a network could be separated into several blocks, a whole architecture is presented as a block set $Arch = \{B^1, B^2, ..., B^n\}$. Each block consists of the above five primitives, which is encoded by a tuple $B^i = (conv, kernel, skip, width, depth)$. Every manipulation for the neural architecture is performed based on the model code.

\subsection{Architecture Scale Search}
\label{subsec:ass}
Most NAS methods~\cite{zoph2017learning,pham2018efficient,liu2017progressive,Real2018Regularized} treat the scale of the architecture as a fixed element based on the prior knowledge from human experts. The scale, the depth and width, of the architecture, usually affects the performance of the architecture. To obtain better performing architectures, we search for the architecture scale by manipulating the width and depth factors.

To accelerate the architecture search process, we employ the parameter sharing method on each model during the search. Inspired by the function-preserving transformations in Net2Net~\cite{chen2015net2net}, namely Net2WiderNet and Net2DeeperNet, we propose a modified parameter sharing method on model training.
When initializing parameters for a network, the proposed algorithm traverses the operation for each layer. If the operation type and the kernel size of the layer consist with that of the shared model, the parameter sharing is applied on this layer. We introduce two parameter sharing behaviors on the width and depth respectively.

\vspace{-12pt}
\paragraph{Parameter sharing on the width-level}
\begin{algorithm}[t]
    \caption{Parameter sharing on the width-level}
    \label{algo:sharewidth}
    \SetKwInOut{Input}{input}\SetKwInOut{Output}{output}
    \Input{kernel $\mathbf{K}_l$ in layer $l$, the original kernel $\mathbf{K}_o$}
    \Output{kernel $\mathbf{K}_l$ in layer $l$}
    \begin{small}
        \texttt{$ch_{in}^s$ $\gets$ min( $ch_{in}^l$, $ch_{in}^o$ ) }
        
        \texttt{$ch_{out}^s$ $\gets$ min($ch_{out}^l$, $ch_{out}^o$)}
        
        \texttt{$\mathbf{K}_l$ $\gets$ $\mathbf{K}_o$($w^o$, $h^o$, $ch_{in}^s$, $ch_{out}^s$)}
    \end{small}
\end{algorithm}

By sharing the parameters, we desire to inherit as more information as possible from the former model. For the convolutional layer, we suppose the convolutional kernel of the $l^{th}$ layer $\mathbf{K}_l$ has the shape of ($w^l$, $h^l$, $ch_{in}^l$, $ch_{out}^l$), where $w^l$ and $h^l$ denote the filter width and height respectively, while $ch_{in}^l$ and $ch_{out}^l$ denote the number of input and output channels respectively. If the original convolutional kernel $\mathbf{K}_o$ has the shape of ($w^o$, $h^o$, $ch_{in}^o$, $ch_{out}^o$), we carry out sharing strategy in Algorithm~\ref{algo:sharewidth}. In addition to the shared parameters, the rest part of $\mathbf{K}_l$ is randomly initialized.

\vspace{-12pt}
\paragraph{Parameter sharing on the depth-level}
The parameters are shared on the depth level in a similar way. Suppose $\mathbf{U}[1, 2, ..., l_u]$ denotes the parameter matrix of one block which has $l_u$ layers, and $\mathbf{W}[1, 2, ..., l_w]$ denotes the parameter matrix of the corresponding block from the shared model which has $l_w$ layers. The sharing process is illustrated in two cases:
\begin{enumerate}[i]
\setlength{\itemsep}{-1pt}
\setlength{\parsep}{0pt}
\setlength{\parskip}{-5pt}
    \item $l_u$ $>$ $l_w$:
    \begin{equation}
        \mathbf{U}[i]=
        \begin{cases}
            \mathbf{W}[i], &\mbox{if}\, i < l_w\\
            \Gamma(i), &\mbox{otherwise}
        \end{cases}
    \end{equation}
    \item $l_u$ $\leq$ $l_w$:
    \begin{equation}
        \mathbf{U}[1, 2, ..., l_u] = \mathbf{W}[1, 2, ..., l_u]
    \end{equation}
\end{enumerate}
where $\Gamma$ is a random weight initializer based on a normal distribution.

\subsection{Population Quality}
\label{subsec:pop_q}
During the evolution process, we design a criterion \emph{population quality} to evaluate the model population. With the search proceeding, the scores of the models in the population improve. To judge whether the population evolution converges, the variance of model scores needs to be taken into consideration. Merely depending on the mean score of models in the population may cause imprecision, because accuracy gains could derive from both parameters sharing and better model performance.

Therefore, until the objective of the population converges to an optimal solution, the mean score of models should be as high while the variance of model scores as low as possible. This issue could be treated as a Pareto-optimal problem~\cite{deb2014multi}. To approximate the Pareto optimal solution, we utilize a target function, population quality, as follows,
\begin{equation}
Q = score_{mean} \times \left [ \frac{std}{target_{std}} \right ]^{\omega }
\label{eq:pop_quality}
\end{equation}
where $\omega$ is the weight factor defined as follows,
\begin{equation}
\label{eq:omega}
\omega=
\begin{cases}
\alpha, & \mbox{if} \ \  std < target_{std}\\
\beta, & \mbox{otherwise}
\end{cases}
\end{equation}
where $\alpha$ and $\beta$ are hyperparameters for controlling the trade-off between the mean score and the standard deviation of scores.

In Eq.~\ref{eq:pop_quality}, $score_{mean}$ denotes the mean score of models in the population, $std$ denotes the standard deviation of model scores and $target_{std}$ is the target $std$. We set $\alpha$ = $\beta$ = -0.07 to assign the value to $\omega$. After the evolution, we pick up the best-quality population and retrain top-k models.

\subsection{Architecture Perturbation Function}
\label{APF}
\begin{algorithm}[t]
    \caption{Architecture Perturbation Function} 
    \label{algo:apf}
    \SetKwInOut{Input}{input}\SetKwInOut{Output}{output}
    \Input{basic architecture $Arch_b$, search space $\mathbb{S}$, number of blocks $N_{\text{blocks}}$, primitives $prims$}
    \Output{perturbed architecture $Arch_p$}
    \begin{small}
        \texttt{$Arch_p$ $\gets$ copy($Arch_b$)}

        \For{\emph{$j$ $<$ $N_{\text{blocks}}$ }}
        {
            $prim$ \texttt{ $\gets$ rand-select(}$prims$\texttt{)}
            
            $value$ \texttt{ $\gets$ rand-generate(}$prim$, $\mathbb{S}$\texttt{)}
            
            $B^j_t$ \texttt{ $\gets$ get-block(}$Arch_t$, j\texttt{)}
            
            $B^j_t[$type$]$ $\gets$ value
        }
    \end{small}
\end{algorithm}

\begin{figure*}[ht]
    \setlength{\abovecaptionskip}{0pt} 
    \setlength{\belowcaptionskip}{-10pt}
    \begin{center}
        \includegraphics[width=0.9\linewidth]{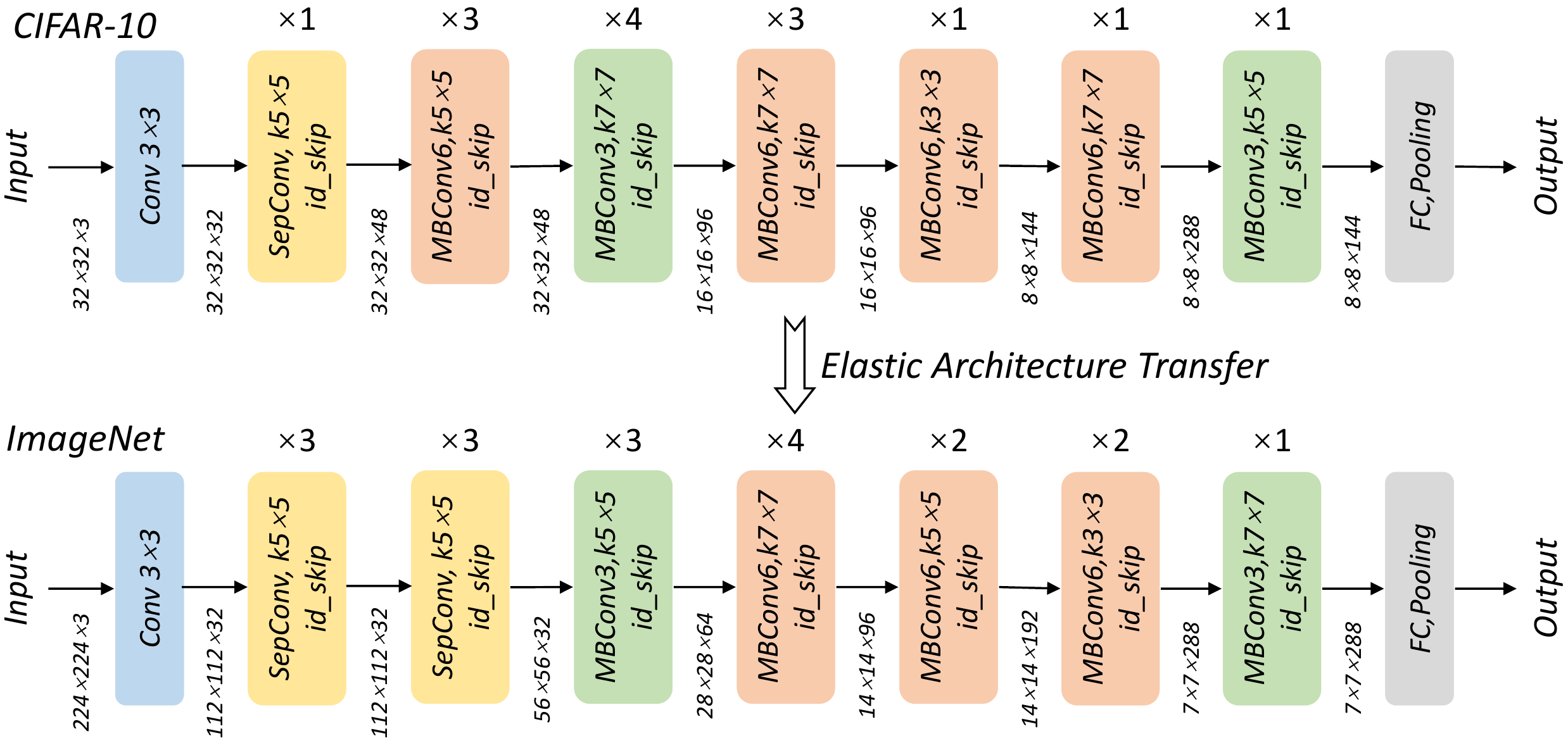}
    \end{center}
    \caption{The architectures searched by EAT-NAS. The upper one is the basic architecture searched on CIFAR-10.
    And the nether one is the architecture EATNet-A searched on ImageNet which is transferred from the basic architecture. 
    }
    \label{fig:Architectures}
\end{figure*}

To transfer the architecture, we initialize the new population on the large scale dataset with the the basic architecture searched on the small dataset as the seed. We design an \emph{architecture perturbation function} to derive new architectures by adding some perturbation to the input architecture code homogeneously and slightly. Algorithm~\ref{algo:apf} illustrates the process of the perturbation function. In each block of the architecture, there are a total of five architecture primitives $(conv, kernel, skip, width, depth)$ to be manipulated as described in Section~\ref{subsec:ss}. We randomly select one type of the five primitives to perturb. For the selected primitive, we generate a new value of the primitive stochastically within the restriction of our search space and replace the existing one. 

When initializing the population on the large dataset, we produce every new architecture by applying the architecture perturbation function to the basic architecture until the number meets the population size. In another word, each initial architecture of the new population is a deformed representation of the basic one. After initializing the new population, the evolution starts the same procedure as described in Algorithm~\ref{algo:ea}. Moreover, the architecture transformation function is utilized as the mutation operation in evolution.

\section{Experiments}
Our experiments mainly consist of two stages, searching for the basic architecture on CIFAR-10 and then transferring it to ImageNet. In this section, we introduce some implementation details in EAT-NAS and report the experimental results. We analyze the results of some ablation experiments which demonstrate the effectiveness of the proposed EAT-NAS method. Other more implementation details are displayed in the appendix of the Supplementary Materials.

\begin{table*}[!hbt]
    \centering
    \caption{ImageNet classification results in the mobile setting. The results of manual-design models are in the top section, other NAS results are presented in the middle section, and the result of our models are in the bottom section.} \label{tab:imagenet results}
    \resizebox{\textwidth}{!}{
    \begin{threeparttable}
    \begin{tabular}{ l  c  c  c c c}
    \toprule
        \textbf{Model} & \textbf{
        \begin{tabular}{c}
             \#Params\\
             (M)
        \end{tabular}
        } & \textbf{
        \begin{tabular}{c}
             \#Mult-Adds\\
             (M)
        \end{tabular}
        } & \textbf{
        \begin{tabular}{c}
             Top-1/Top-5\\
             Acc(\%)
        \end{tabular}
        } &\textbf{Search Dataset} &\textbf{
              \begin{tabular}{c}
                  Search Time\\
                  (GPU hours)
              \end{tabular}
              }\\ 
        \midrule
MobileNet-v1~\cite{howard2017mobilenets} & 4.2 & 575 & 70.6 / 89.5 & - & - \\ 
MobileNet-v2~\cite{sandler2018mobilenetv2} & 3.4 & 300 & 71.7 / -$\;\;\;\;\;$ & - & -\\
MobileNet-v2 (1.4)\cite{sandler2018inverted} & 6.9 & 585 & 74.7 / -$\;\;\;\;\;$ & - & -\\
ShuffleNet-v1 2x \cite{zhang2017shufflenet} & $ \approx $ 5 & 524 & 73.7 / -$\;\;\;\;\;$ & - & -\\
        \midrule
NASNet-A~\cite{zoph2017learning} & 5.3 & 564 & 74.0 / 91.6 & CIFAR-10 & 48K\\ 
NASNet-B~\cite{zoph2017learning} & 5.3 & 488 & 72.8 / 91.3 & CIFAR-10 & 48K\\
NASNet-C~\cite{zoph2017learning} & 4.9 & 558 & 72.5 / 91.0 & CIFAR-10 & 48K\\
AmoebaNet-A~\cite{Real2018Regularized} & 5.1 & 555 & 74.5 / 92.0 & CIFAR-10 & 76K\\
AmoebaNet-B~\cite{Real2018Regularized} & 5.3 & 555 & 74.0 / 91.5 & CIFAR-10 & 76K\\
AmoebaNet-C~\cite{Real2018Regularized} & 5.1 & 535 & 75.1 / 92.1 & CIFAR-10 & 76K\\
PNASNet-5~\cite{liu2017progressive} & 5.1 & 588 & 74.2 / 91.9 & CIFAR-10 & 6K\\
MnasNet~\cite{2018arXiv180711626T} & 4.2 & 317 & 74.0 / 91.8 & ImageNet & 91K\\ 
MnasNet (our impl.) & 4.2 & 317 & 73.3 / 91.3 & ImageNet & 91K\\
DARTS~\cite{liu2018darts} & 4.7 & 574 & 73.3 / 91.3 & CIFAR-10 & 96\\
SNAS~\cite{xie2018snas} & 4.3 & 522 & 72.7 / 90.8 & CIFAR-10 & 36 \\
        \midrule
EATNet-A & 5.1 & 563 & 74.7 / 92.0 & CIFAR-10 to ImageNet & 856\\ 
EATNet-B & 5.3 & 551 & 74.2 / 91.8 & CIFAR-10 to ImageNet & 856\\ 
EATNet-S & 4.6 & 414 & 72.7 / 91.1 & CIFAR-10 to ImageNet & 856\\ 
        \bottomrule
    \end{tabular}
    
    \begin{tablenotes}
        \item[*] To avoid any discrepancy between different implementations or training settings, we try our best to reproduce the performance of MnasNet~\cite{2018arXiv180711626T}. The highest accuracy we could reproduce is 73.3\% for MnasNet. All the training settings we use are the same as that we reproduce MnasNet for the fair comparison.
    \end{tablenotes}
  \end{threeparttable}
    }
\vspace{-15pt}
\end{table*}

\subsection{Search on CIFAR-10}

The experiments on CIFAR-10 are divided into two steps including architecture search and architecture evaluation. CIFAR-10 consists of 50,000 training images and 10,000 testing images. We split the original training set (80\% - 20\%) to create our training and validation sets for the search process. The original CIFAR-10 testing set is only utilized in the evaluation process of the final searched models. All images are whitened with the channel mean subtracted and the channel standard deviation divided. Then we crop 32 $\times$ 32 patches from images padded to 40 x 40 and randomly flip them horizontally.

During the search process, we set the population size as 64 and the sample size as 16. Every model generated during evolution is trained for 1 epoch and is evaluated on the separate validation set. We mutate about 1,400 models during the total evolution. The number of model parameters is the sub-optimizing objective with the target as 3.0M. Each model on CIFAR-10 consists of 7 blocks and the downsampling operations are carried out in the third and fifth blocks. The initial number of channels is 32. For evaluation, we retrain top-8 models searched on CIFAR-10 and select the one with the best accuracy as the basic model. Since the CIFAR-10 results are subject to high variance even with exactly the same setup~\cite{liu2017hierarchical}, we report the mean and standard deviation of 5 independent runs for our model. The basic model achieves 96.42\% mean test accuracy (the standard deviation of 0.05) with only 2.04M parameters. The architecture of the basic model is shown in Fig.~\ref{fig:Architectures}.

\subsection{Transferring to ImageNet}

We use the architecture of the basic model searched on CIFAR-10 as the seed to generate models of the population on ImageNet. The search process is carried out on the whole ImageNet training dataset. To avoid overfitting the original ImageNet validation set, we have a separate validation set which contains 50K images randomly selected from the training set to measure the accuracy. We use the architecture perturbation function to produce 64 new architectures based on basic architecture. 
During architecture search, we train every model for one epoch. The number of multiply-add operations is set as the sub-optimizing objective during evolution with the target as 500M. The input resolution of the network is set to $224\times224$. Each model is composed of 7 blocks and the number of input channels is 32 as well. The downsampling operations are carried out in the input layer and the 2nd, 3rd, 4th, 6th block.

\begin{figure*}[htbp]
    \setlength{\abovecaptionskip}{5pt} 
    \setlength{\belowcaptionskip}{-10pt}
    \centering
        \subfloat[The mean accuracy of the models in the population.]{
            \includegraphics[width=0.35\textwidth]{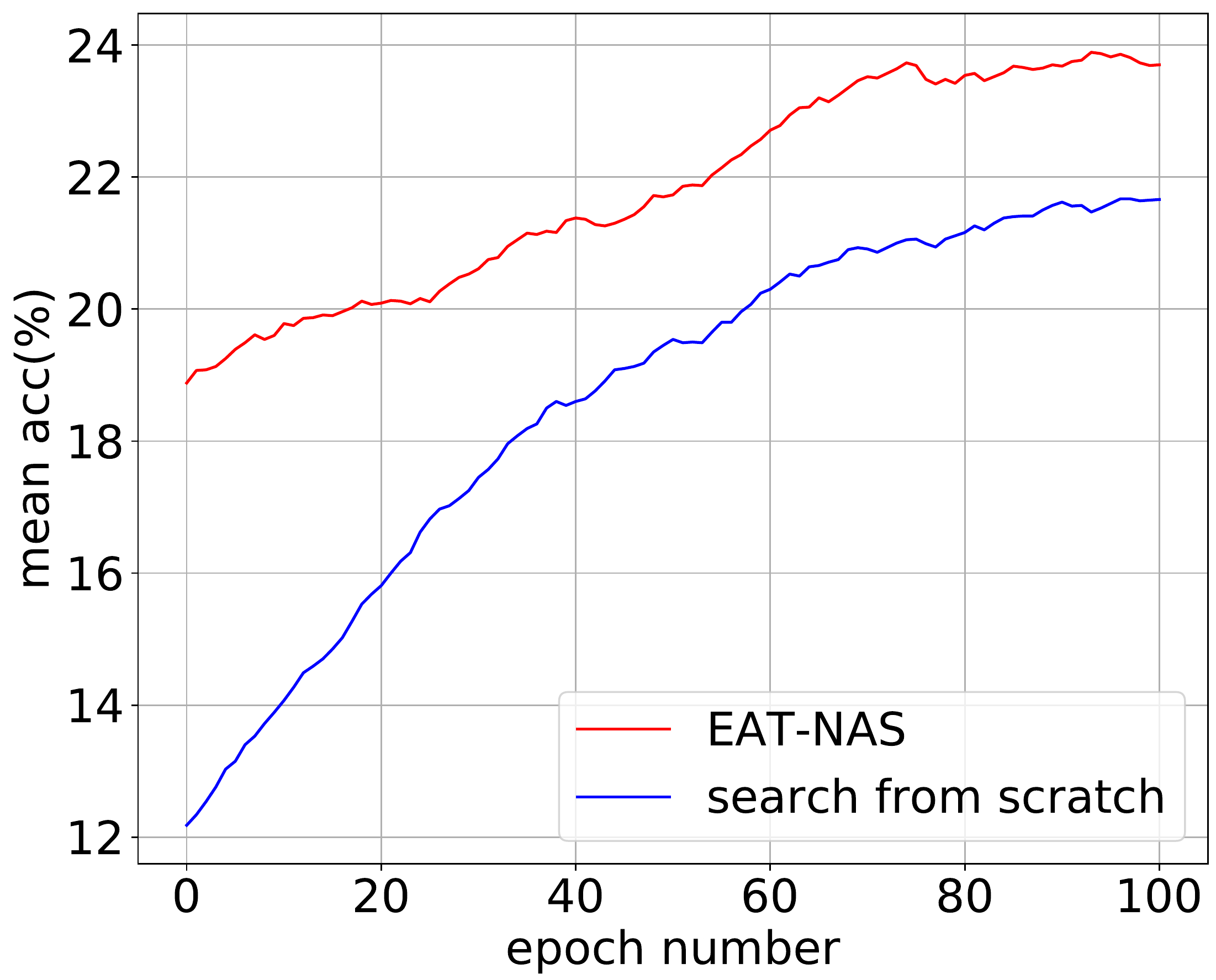}
        } \qquad
        \subfloat[The population quality.]{
            \includegraphics[width=0.35\textwidth]{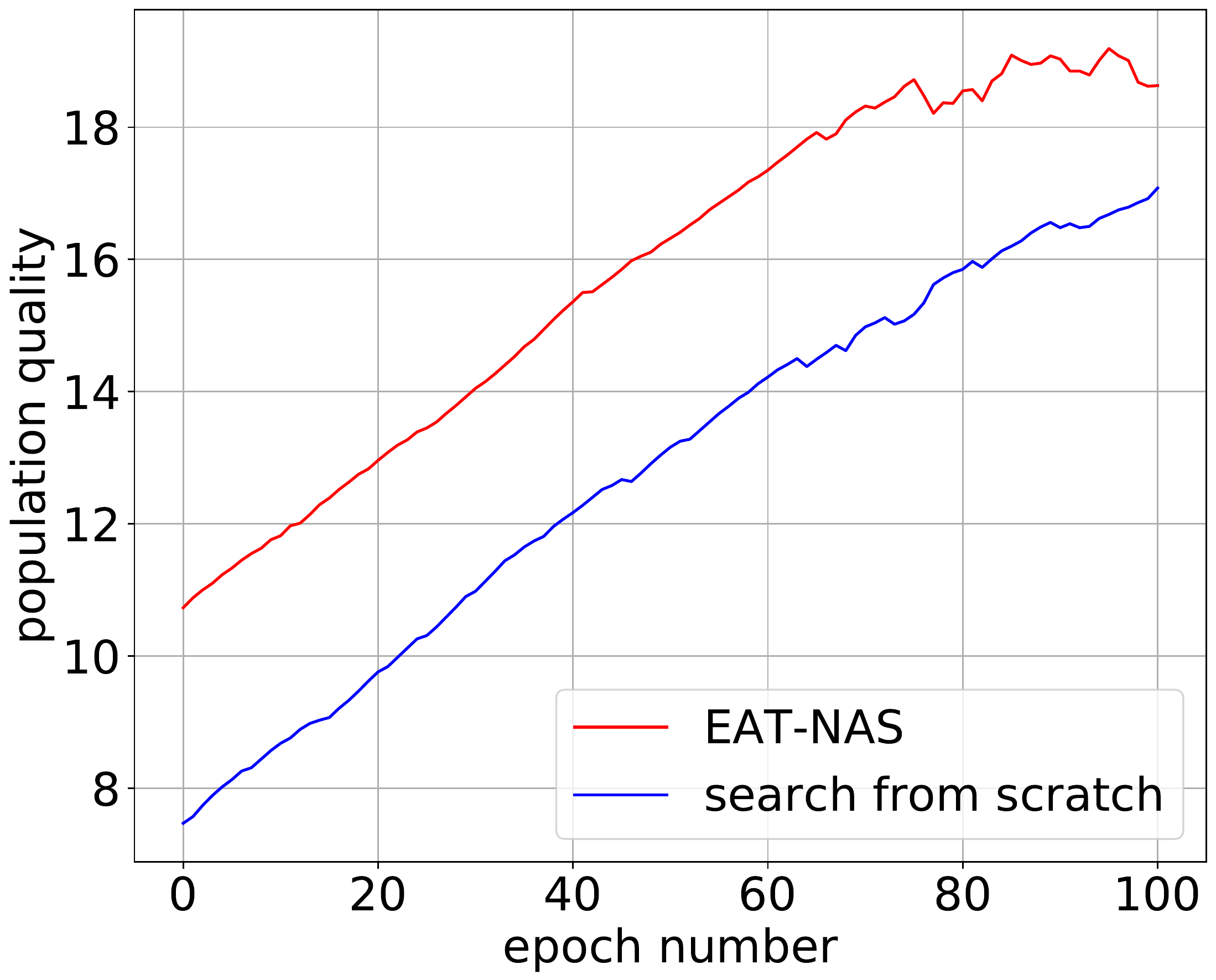}
        }
    \caption{Comparing the evolution process on ImageNet of EAT-NAS and search from scratch.}
    \label{fig:compare}
\end{figure*}

As shown in Fig.~\ref{fig:compare}, the evolution process takes about 100 evolution epochs to converge. In another word, taking the initial 64 models into account, we only sample around 164 models to find out the best one based on the basic architecture. In MnasNet~\cite{2018arXiv180711626T}, the controller samples about 8K models during architecture search, 50 times the amount of ours. With much less computational resources, EAT-NAS achieves comparable results on ImageNet as in Table~\ref{tab:imagenet results}.

Fig.~\ref{fig:Architectures} displays the basic architecture and the architecture of EATNet-A. From the figure, we find that though based on the basic architecture, there are some transformations on EATNet-A. During the elastic architecture transfer process, all architecture primitives in the basic architecture are likely to be modified. For example, the operation type in the second block has changed from \emph{Mbconv6} to \emph{SepConv}, and the kernel sizes have changed in some blocks. The depth and the width of the architecture have changed as well. The modifications to the architecture primitives adapt the architecture to the new dataset.

In summary, our EAT-NAS includes two stages, search on CIFAR-10 and transfer to ImageNet. It takes 22 hours on 4 GPUs to search for the basic architecture on CIFAR-10 and 4 days on 8 GPUs to transfer to ImageNet. Though DARTS~\cite{liu2018darts} and SNAS~\cite{xie2018snas} take less search time, they only search on the small dataset, CIFAR-10. And our ImageNet performances surpass them clearly.

\begin{figure}[ht]
    \setlength{\abovecaptionskip}{-5pt} 
    \setlength{\belowcaptionskip}{0pt}
    \begin{center}
        \includegraphics[width=0.8\linewidth]{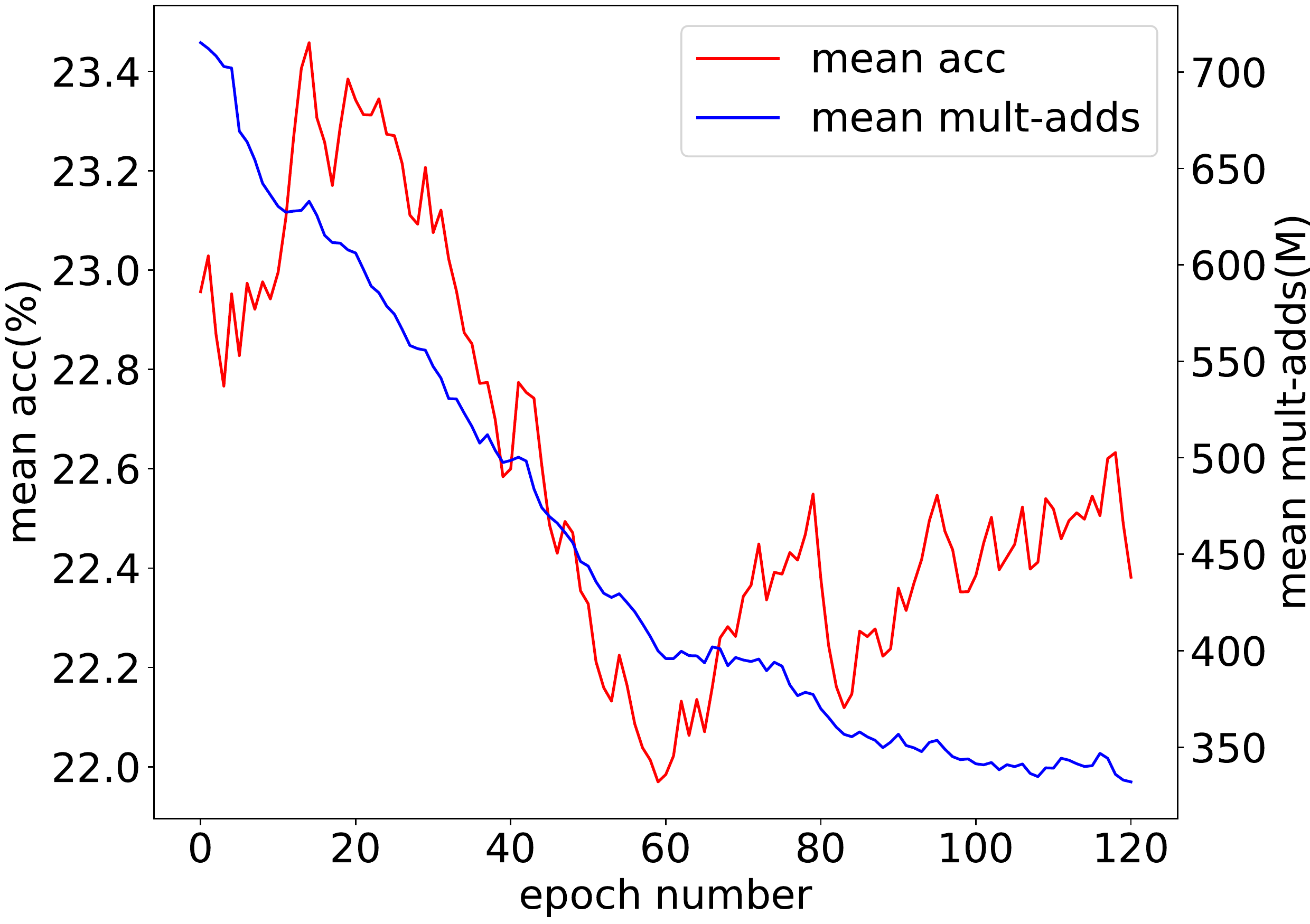}
    \end{center}
    \caption{The mean accuracy and the mean multiply-adds of the models during the search on ImageNet whose basic architecture has worse performance on CIFAR-10.
    }
    \label{fig:bad-basic}
\end{figure}

\subsection{Ablation Study}
\label{subsec:ablation-study}

\paragraph{Efficiency of EAT}

To demonstrate the efficiency of our proposed method EAT-NAS, we carry out the search process on ImageNet from scratch. All the settings are the same as EAT-NAS in both search and evaluation process. The search process takes the same GPU hours as well. Fig.~\ref{fig:compare} shows the mean accuracy of models in the population and the population quality of EAT-NAS and search from scratch on ImageNet for equal search epochs. We find out that after initialization, the mean accuracy of models is obviously higher of EAT-NAS than that of search from scratch all through the search process. And the evolution process converges faster for EAT-NAS. In Table~\ref{tab:contrast results}, we compare the best performing model of top-8 searched from scratch with that from EAT-NAS. The compared models we select are guaranteed to have similar model sizes. The model EATNet-S surpasses that searched from scratch obviously.

\vspace{-14pt}
\paragraph{Effectiveness of EAT}

To verify the effectiveness of our elastic architecture transfer method, we apply our basic architecture searched on CIFAR-10 directly on ImageNet without any modification as the handcrafted transfer does. We train the basic model on ImageNet under the same settings of EAT-NAS. The performance of the basic model is shown in Table~\ref{tab:contrast results} as well. The basic model achieved a high validation accuracy but with a very large number of multiply-add operations. Its high validation accuracy demonstrates the superior quality of the initialization seed in the transfer process. Comparing EATNet-A with Basic Model, we find that not only does EAT promote the accuracy of the model, but also optimizes the size of the model.

\vspace{-14pt}
\paragraph{Impact of Basic Architecture Performance}
We select one architecture with worse performance as the basic architecture in our transfer process, whose validation accuracy on CIFAR-10 is 96.16\% and the number of parameters is 1.9M. As Fig.~\ref{fig:bad-basic} shows, the basic architecture with worse performance has a negative impact on transfer. The mean accuracy degrades in the preliminary epochs. This experiment demonstrates the importance of a well-performing basic architecture for transfer. We retrain the searched top-8 models under the same settings and compare the best one with that searched by EAT-NAS which has a similar model size. As Table~\ref{tab:contrast results} displays, models searched by EAT-NAS surpass that searched with the bad-performing basic architecture. We attribute the results to the performance of the basic architecture.

\begin{table}[t]
    \setlength{\abovecaptionskip}{0pt} 
    \centering
    \caption{The results of contrast experiments on ImageNet. SS denotes the model searched from scratch on ImageNet. The basic model searched on CIFAR-10 is directly applied on ImageNet without any modification. Model-B denotes the best model searched on ImageNet with a worse-performing basic architecture.}
    \label{tab:contrast results}
    \resizebox{0.475\textwidth}{!}{
        \begin{threeparttable}
            \begin{tabular}{ l  c  c  c}
                \toprule
                \textbf{Model} & \textbf{
                \begin{tabular}{c}
                     \#Params\\(M)
                \end{tabular}
                } & \textbf{
                \begin{tabular}{c}
                     \#Mult-Adds\\(M)
                \end{tabular}
                } & \textbf{
                \begin{tabular}{c}
                     Top-1/Top-5\\
                     Acc (\%)
                \end{tabular}
                }\\ 
                \midrule
                SS & 5.53 & 447 & 71.8 / 90.6\\
                \midrule
                Basic Model & 3.14 & 886 & 74.3 / 92.0\\
                \midrule
                Model-B & 3.20 & 405 & 71.7 / 90.5\\
                \midrule
                EATNet-A & 5.08 & 563 & 74.7 / 92.0\\
                EATNet-B & 5.25 & 551 & 74.2 / 91.8\\
                EATNet-S & 4.63 & 414 & 72.7 / 91.1\\
                \bottomrule
            \end{tabular}
        \end{threeparttable}
    }
\vspace{2pt}
\end{table} 

\vspace{-10pt}
\section{Conclusion and Future Work}
In this paper, we propose an elastic architecture transfer mechanism for accelerating the large-scale neural architecture search (EAT-NAS). Rather than spending a lot of computation resources to directly search the neural architectures on large-scale tasks, EAT-NAS makes full use of the information of the basic architecture searched on the small-scale task. We transfer the basic architecture with elasticity to the large-scale task fast and precisely. With less computational resources, we obtain networks with excellent ImageNet classification results in mobile sizes.

In the future, we would try to combine the proposed mechanism with other search methods, such as reinforcement learning and gradient-based NAS. In addition, EAT-NAS can also be utilized to search for neural architectures in other computer vision tasks like detection, segmentation, and tracking, which we also leave for future work.

\section*{Acknowledgement}
We thank Liangchen Song and Guoli Wang for the discussion and assistance.

{\small
\bibliographystyle{ieee}
\bibliography{egbib}
}

\begin{appendices}

\section{Experiment Details}
We provide more experiment details in this appendix. Our experiments mainly consist of two stages, searching for the basic architecture on CIFAR-10 and then transferring it to ImageNet.

\subsection{Search on CIFAR-10}

For each model during the search ,the depth and width of the mutated model would vary in an extremely wide range within our search space if there is not any restriction. Some constraints are added to the scale of the model within an acceptable range to avoid the memory running out of control during the search. On CIFAR-10, the total expansion ratio is limited within $[4, 10]$.

For training during the search process, the batch size is set as 128. We use SGD optimizer with the learning rate of 0.0256 (fixed during the search), momentum 0.9, and weight decay 3 $\times$ $10^{-4}$. The search experiment is carried out on 4 GPUs which takes about 22 hours. For evaluation, every model is trained for 630 epochs with a batch size of 96. The initial learning rate is 0.0125 and the learning rate follows the cosine annealing restart schedule. Other hyperparameters remain the same as that in the search process. Following existing works~\cite{pham2018efficient, zoph2017learning, liu2017progressive, Real2018Regularized}, additional enhancements include cutout~\cite{devries2017improved} with the length of 16, and auxiliary towers with weight 0.4. The training of the searched model takes around 13 hours on two GPUs.

\subsection{Transferring to ImageNet}

During architecture search, we train every model for one epoch with batch 128 and a learning rate of 0.05. Following GoogLeNet~\cite{DBLP:conf/cvpr/SzegedyLJSRAEVR15}, the input images are sampled as various sized patches of the image whose size is distributed between 20\% and 100\% of the image area with aspect ratio constrained to the interval $[\frac{3}{4}, \frac{4}{3}]$. The input resolution of the network is set to $224\times224$. Other hyperparameters of the search process are the same as that on CIFAR-10. For each model, the number of layers is limited within $[16, 18]$ and the total width expansion ratio is limited within $[8, 16]$.

For evaluating the model performance on ImageNet, we retrain the final top-8 models on $224 \times 224$ images of the training dataset for 200 epochs, using standard SGD optimizer with momentum rate set to 0.9, weight decay $4 \times 10^{-5}$ and 0.1-weighted label smoothing. Our batch size is 256 on 4 GPUs. The initial learning rate is 0.1 and it decays in a polynomial schedule to $1 \times 10^{-4}$. We follow ResNet~\cite{he2016deep} to do the data augmentation. We resize the original input images with its shorter side randomly sampled in [256, 480] and $1 \pm \frac{1}{4}$ aspect ratio for scale augmentation~\cite{simonyan2014very}. A $224 \times 224$ patch is randomly cropped from an image or its horizontal flip, with the per-pixel mean subtracted~\cite{krizhevsky2012imagenet}. For the last 20 epochs, we only keep the $256 \times N$ for the resizing scale to fine-tune the model. To avoid any discrepancy between different implementations or training settings, we try our best to reproduce the performance of MnasNet (74.0\% on ImageNet). The highest accuracy we could reproduce is 73.3\% for MnasNet. All the training settings we use are the same as that we reproduce MnasNet for the fair comparison.

\end{appendices}

\end{document}